%% file: 1372_cameraready.tex
\documentclass[twoside]{article}

\usepackage[accepted]{aistats2022}
\usepackage[T1]{fontenc}
\usepackage{lmodern}
\usepackage{epsfig}
\usepackage{float}
\usepackage{afterpage}
\usepackage{amsmath}
\usepackage{amstext}
\usepackage{amssymb,bm}
\usepackage{latexsym}
\usepackage{color}
\usepackage{graphicx}
\usepackage{amsmath}
\usepackage{amsthm}
\usepackage{graphicx}
\usepackage[center]{caption}
\usepackage{pstricks}
\usepackage{caption}
\usepackage{subcaption}
\usepackage{booktabs}
\usepackage{multicol}
\usepackage{lipsum}% dummy text
\usepackage{mathtools} %\coloneqq for perfect :=
\usepackage{enumitem}

\usepackage{tikz}
\usetikzlibrary{patterns,perspective}
\usepackage{pgfplots}
\pgfplotsset{compat=1.12}

\newtheorem{remark}{Remark}

\newtheorem{prop}{Proposition}

\newtheorem{definition}{Definition}
\newtheorem{theorem}{Theorem}

\newtheorem{example}{Example} 

\newtheorem{assumption}{Assumption}

\newcommand{\eu}{\mathrm{e}}

\newcommand{\E}{\mathbb{E}}

\newcommand{\ex}{\mathop{\mathrm{ex}}\nolimits}
\DeclarePairedDelimiter\parentheses{\lparen}{\rparen}

\makeatletter
\newcommand*{\transpose}{%
	{\mathpalette\@transpose{}}%
}
\newcommand*{\@transpose}[2]{%
	\raisebox{\depth}{$\m@th#1\intercal$}%
}
\makeatother

\begin{document}

\runningtitle{Dimensionality Reduction for Finding Least Favorable Priors with a Focus on Bregman Divergence}


\twocolumn[

\aistatstitle{A Dimensionality Reduction Method for Finding Least Favorable Priors with a Focus on Bregman Divergence}

\aistatsauthor{Alex Dytso \And Mario Goldenbaum \And  H. Vincent Poor \And Shlomo Shamai (Shitz) }

\aistatsaddress{New Jersey Institute\\ of Technology \And  Bremen University of\\ Applied Sciences \And Princeton University \And Technion - Israel\\ Institute of Technology}]

\begin{abstract}
A common way of characterizing minimax estimators in point estimation is by moving the problem into the Bayesian estimation domain and finding a least favorable prior distribution. The Bayesian estimator induced by a least favorable prior, under mild conditions, is then known to be minimax. However, finding least favorable distributions can be challenging due to inherent optimization over the space of probability distributions, which is infinite-dimensional. This paper develops a dimensionality reduction method that allows us to move the optimization to a finite-dimensional setting with an explicit bound on the dimension. The benefit of this dimensionality reduction is that it permits the use of popular algorithms such as projected gradient ascent to find least favorable priors. Throughout the paper, in order to make progress on the problem, we restrict ourselves to Bayesian risks induced by a relatively large class of loss functions, namely Bregman divergences.
\end{abstract}%

%%%%%%%%%%%%%%%%%%%%%%%%%%%%%%%%%%%%%%%%%%%%%%%%%%%%%%%%%%%%%%%%%%%%%%%%%%%%%%%%%%%%%%%%%%%%%%%%%%%%%%%%%
\section{INTRODUCTION}
\label{sec:Introduction}
Consider the problem of estimating a deterministic parameter $x \in \Omega \subseteq \mathbb{R}^n$ from a noisy observation $Y \in \mathcal{Y} \subseteq \mathbb{R}^k$, where $x$ and $Y$ are related through a conditional distribution $P_{Y|X=x}$. The standard objective in estimation theory is to find an estimator $f(Y)$ that minimizes some risk function $r(x,f)$. Formally, the risk function can be defined as 
\begin{equation}
	r(x,f)\coloneqq\E\bigl[\ell\bigl(x,f(Y)\bigr)\bigr], 
\end{equation}% 
where the expectation is taken with respect to $P_{Y|X=x}$ and 
where $\ell:\Omega \times \mathbb{R}^n \to [0,\infty)$ is some loss function (e.g., square loss). A common design principle is to look for an estimator that achieves the smallest maximum risk among all estimators. Such an estimator is called \emph{minimax}. More precisely, an estimator $f_M: \mathcal{Y} \to \mathbb{R}^n$ is said to be minimax if 
\begin{equation}
	\sup_{x \in \Omega} r(x,f_M)=  \inf_{f} \sup_{x \in \Omega} r(x,f), 
\end{equation}% 
 where the infimum is taken over all measurable functions.
%
%{\color{red}[MG] The range of $f_M$ is $\mathbb{R}^k$ while in the definition of loss function $\ell$ the range of $f$ seems to be $\mathbb{R}^m$. Different dimensions?}

%{\color{red}[MG] General question. When taking in (2) the inf over $f$, do we choose $f$ from the space of ``all'' functions $\mathbb{R}^k\to\mathbb{R}^m$ or de we have any restrictions on the choice of function?}

A potential way of finding the minimax estimator is to go along the Bayesian route. More specifically, consider the problem of estimating a random vector $X\in\Omega\subseteq\mathbb{R}^n$ with prior distribution $P_X$ from a noisy observation $Y \in \mathcal{Y} \subseteq \mathbb{R}^k$ that are related through the same conditional distribution $P_{Y|X}$. The minimum Bayesian risk for prior $P_X$ is then defined as
\begin{equation}
	R(P_X, P_{Y|X})\coloneqq\inf_{f}\E\bigl[r(X,f)\bigr],
\end{equation}
where the expectation is taken with respect to $P_X$. 
Furthermore, a prior distribution $P_{X}^\star$ is said to be \emph{least favorable} if  
\begin{equation}
	R(P_X^\star, P_{Y|X}) = \sup_{P_X}R(P_X, P_{Y|X}), 
	\label{eq:Main_Optimization} 
\end{equation}%
where the supremum is taken over all distributions supported on $\Omega$. In other words, the random vector that follows $P_X^\star$ is the `hardest' to estimate. A classical result in estimation theory states that 
\begin{equation}
	f_{P_X^\star}  \in \arg \ \inf_{f} \E_{P_X^\star}\bigl[  r(X,f )\bigr]; 
\end{equation}%
that is, a best estimator for the least favorable prior is  also a minimax estimator. See for instance \cite{lehmann2006theory} where this is shown under very mild conditions. Due to this fact, finding least favorable prior distributions has received considerable attention in the literature. 

%{\color{red}[MG] Is the expectation in (5) with respect to $P_X$ or $P_X^*$?}  {\color{blue} $P_X^\star$ } 

However, finding least favorable priors is a formidable task. The difficulty stems from the fact that the optimization in \eqref{eq:Main_Optimization} is done over the space of probability distributions, which results in an infinite-dimensional optimization problem. The objective of this work is to show that under mild conditions the optimization problem in \eqref{eq:Main_Optimization} can be reduced to a finite dimensional one. The key benefit of such a reduction is that one can begin to use numerical recipes to find a least favorable prior (e.g., gradient ascent algorithm).  

The outline and the contributions of the paper are as follows. The remaining part of Section~\ref{sec:Introduction} is dedicated to notational remarks and past work. Section~\ref{sec:prelim} presents some preliminary definitions (e.g., Bregman divergence), provides the problem statement and discusses our assumptions. Section~\ref{sec:Main_Result} presents our main results, which show that under certain general conditions finding a least favorable prior can be reduced to a finite-dimensional optimization problem. Section~\ref{sec:Proof_main} is devoted to the proof of the main theorem. Section~\ref{sec:Simulations} builds on the results of Section~\ref{sec:Main_Result} and  discusses how a  projected gradient ascent algorithm can be used to find least favorable prior distributions. The algorithm is then applied to find least favorable priors in the context of binomial noise and quantized Gaussian noise.
%
%
%
%%%%%%%%%%%%%%%%%%%%%%%%%%%%%%%%%%%%%%%%%%%%%%%%%%%%%%%%%%%%%%%%%%%%%%%%%%%%%%%%%%%%%%%%%%%%%%%%%%%%%%%%%%%%%%%%%%%
\subsection{Notational Remarks}\label{sec:notation}%
Deterministic scalars and vectors are denoted by lower case letters and random objects by capital letters; $\mathbb{R}$ denotes the affinely extended real number system;  $\|\cdot\|$ denotes the Euclidean norm; the closed ball in $\mathbb{R}^n$ of radius $r$ centered at $x$ is denoted as $\mathcal{B}_x(r)\coloneqq\{y\in\mathbb{R}^n:\|y-x\| \le r \}$;  for a random vector $X \in \mathbb{R}^n$ and every measurable set $\mathcal{A} \subset \mathbb{R}^n$ we denote the probability measure of $X$ as $P_X(\mathcal{A})=\mathbb{P}[X\in\mathcal{A}]$; if it is clear from the context, we sometimes write $P$ instead of $P_X$; the space of all probability measures defined on sample space $\Omega\subseteq\mathbb{R}^n$ is denoted as $\mathcal{P}(\Omega)$; the Dirac measure centered on a fixed point $x$ is denoted as $\delta_x$; for two probability distributions $P$ and $Q$, $P \ll Q$ means $P$ is absolutely continuous with respect to $Q$; for a random vector $X$ with distribution $P_X$ the expected value is $\E[ X ]= \int x\,\mathrm{d}P_X(x)$ and when we need to emphasize that $X$ is distributed according to $P_X$ we use $\E_{P_X}[X]$. 
%
%
%
%
%%%%%%%%%%%%%%%%%%%%%%%%%%%%%%%%%%%%%%%%%%%%%%%%%%%%%%%%%%%%%%%%%%%%%%%%%%%%%%%%%%%%%%%%%%%%%%%%%%%%%%%%%%%%%%%%%%%
\subsection{Past Work}
The theory of finding least favorable prior distributions has received some attention for the special case when the noise is Gaussian and  the loss function quadratic, for which $R( P_X, P_{Y|X})$ is commonly known as the \emph{minimum mean square error}. For the univariate case (i.e., $n=1$), Ghosh has shown in \cite{ghosh1964uniform} that if the support of $X$ is bounded and the noise is Gaussian (i.e, $P_{Y|X}=\mathcal{N}(x,1)$), then least favorable priors are discrete with finitely many mass points. 
Also for $P_{Y|X}=\mathcal{N}(x,1)$, the authors of \cite{casella1981estimating} capitalized on the result of Ghosh and provided necessary and sufficient conditions for the optimality of a two mass points prior distribution, and sufficient conditions for the optimality of three mass points priors. In \cite{berry1990minimax}, Berry has extended the results of \cite{casella1981estimating} to the case of multivariate Gaussian noise with covariance taken to be the identity matrix. For $n\geq 1$, the authors of \cite{dytsoISIT2018} have considered generalized moment constraints or linear constraints (i.e., $\E[g(X)] \le c$) on $X$, and have shown that if $g\in o(\|x\|^2)$, then the support of a least favorable distribution is unbounded, and if $g\in\omega(\|x\|^2)$ it is bounded.  
 
Much less work has been done for the general case. In \cite{kempthorne1987numerical}, for instance, it has been shown for $n=1$ that if the conditional Bayes risk $\E \left[  \ell \left(X, f(Y)\right) |X=x \right]$ is an analytic function of $x$ and the support is bounded, then the least favorable prior is discrete with finitely many mass points. For a summary of known results on the properties of least favorable priors together with some extensions the interested reader is referred to \cite{marchand2004estimation}. 
 
Algorithms for computing least favorable priors have been proposed in \cite{kempthorne1987numerical, nelson1966minimax} and have been shown to converge under certain conditions. However, as these algorithms were designed without an explicit upper bound on the number of mass points in the support, their procedure relies on an optimization over an infinite dimensional space. Furthermore, it is not difficult to show that the cutting-plane algorithm, proposed in \cite{huang2005characterization} for finding the capacity-achieving distribution of a communication channel, can be adapted to the setting of finding a least favorable prior. However, it shares the same drawback as the algorithms proposed in \cite{kempthorne1987numerical, nelson1966minimax} as it also relies on an optimization over an infinite dimensional space.  

Finally, our method is inspired by the dimensionality reduction studied in the context of mutual information in \cite{witsenhausen1980convexity}, from which we borrow several key ideas (e.g., Dubins' theorem). See also \cite{Dytso:Goldenbaum:Poor:Shamai:CISS2018}. 
% Bregman Divergences have been introduced in \cite{bregman1967relaxation} in the context of  convex optimization.  {\red TO DO: give more references} 
% 
% {\red  I will add the reference later.     } 
% 
%
%
%
%%%%%%%%%%%%%%%%%%%%%%%%%%%%%%%%%%%%%%%%%%%%%%%%%%%%%%%%%%%%%%%%%%%%%%%%%%%%%%%%%%%%%%%%%%%%%%%%%%%%%%%%%%%%%%%%%%%%%%% 
\section{PRELIMINARIES AND PROBLEM STATEMENT }
\label{sec:prelim}
%
%
%
%%%%%%%%%%%%%%%%%%%%%%%%%%%%%%%%%%%%%%%%%%%%%%%%%%%%%%%%%%%%%%%%%%%%%%%%%%%%%%%%%%%%%%%%%%%%%%%%%%%%%%%%%%%%%%%%%%%
\subsection{Bregman Divergence and Bayesian Risk}
\label{sec:Bregman Divergence:Basics}
To even have a chance to solve the optimization problem in \eqref{eq:Main_Optimization}, we need to slightly restrict the class of loss functions. To that end, we will consider the following class, which is in fact very large. 
\begin{definition} \emph{(Bregman Divergence)}  
	Let $\phi:\Omega\to\mathbb{R}$ be a \emph{continuously differentiable} and \emph{strictly convex} function. The Bregman divergence associated with $\phi$ is then defined as
	\begin{equation}
		\ell_\phi(u,v) =\phi(u)-\phi(v)-\bigl\langle u-v,\nabla \phi(v)\bigr\rangle. 
	\end{equation}% 
\end{definition}%

%{\color{blue} 
The classical squared error loss is recovered through Bregman divergences by choosing $\phi(u)=\|u\|^2$. As another example, consider the function $\phi(u)=u_1 \log u_1 + u_2 \log u_2$  where $u=[u_1,u_2]^\transpose$ with $\Omega=\mathbb{R}_{+}^2$,  which induces the following  Bregman divergence (known as the generalized I-divergence): for $u=[u_1,u_2]^\transpose$ and $v=[v_1,v_2]^\transpose$ 
\begin{equation}
    \ell_\phi(u,v) =   u_1 \log \frac{u_1}{v_1}+u_2 \log \frac{u_2}{v_2} -(u_1 -v_1)-(u_2 -v_2). 
\end{equation} %

Bregman divergences have been introduced in \cite{bregman1967relaxation} in the context of convex optimization. 
In \cite{csiszar1991least}, Bregman divergences, together with $f$-divergences, were characterized axiomatically and considered in estimation settings. A thorough investigation of their properties was undertaken in \cite{banerjee2005clustering}, where it was shown that many commonly used loss functions are members of this family. The structure of the optimal estimator under Bregman divergences as loss functions was studied in \cite{banerjee2005optimality}, where it was shown that the conditional expectation is the unique minimizer.
For extensions of Bregman divergence to different spaces the interested reader is referred to  \cite{frigyik2008functional,iyer2012submodular,wang2014bregman} and references therein. 

%} 
%
%{\color{red}[MG] Does it make sense to call it ``Bregman loss'' and then mention it is nothing but the Bregman divergence? Just a thought.}
%
%{\color{red}[MG] So in comparison to the loss function in (1), which is $\ell:\Omega\times\mathbb{R}^m\to\mathbb{R}$, Bregman loss is $\ell_\phi:\Omega\times\Omega\to\mathbb{R}$?}

\begin{definition} \emph{(Bayesian Risk with Respect to Bregman Divergence.)} 
	For a joint distribution $P_{XY}$, we denote the Bayesian risk with respect to loss function $\ell_\phi$ as 
	\begin{equation}
		R_\phi( P_X, P_{Y|X})\coloneqq\inf_{f: f\textup{ is measurable}}\E\bigl[\ell_\phi\bigl(X,f(Y)\bigr)\bigr]. 
	\end{equation}%
\end{definition}% 

The following theorem summarizes some fundamental properties of $\ell_\phi$ and $R_\phi(P_X, P_{Y|X})$, the proof of which can be found in \cite{banerjee2005clustering} and \cite{banerjee2005optimality}.
\begin{theorem} \emph{(Fundamental Properties of Bregman Divergence and Bayesian Risk)}
	\begin{enumerate}
		\item \emph{(Non-Negativity)}  $\forall u,v\in\Omega:\ell_\phi(u,v) \ge 0$, with equality if and only if $u=v$;
		\item \emph{(Convexity)}  $\ell_\phi(u,v)$ is convex in $u$;
		\item  \emph{(Linearity)}  $\ell_\phi(u,v)$ is  linear in $\phi$;
		\item \emph{(Orthogonality Principle and Pythagorean Identity)}  For every random variable $X\in \Omega$ and every $u\in \Omega$ 
			\begin{equation}
				\E\bigl[\ell_\phi(X,u)\bigr]=\E\bigl[\ell_\phi\bigl(X,\E[X]\bigr)\bigr]+\ell_\phi\bigl(\E[X],u\bigr).
			\end{equation}%
			Moreover, for any $f(Y)$
			\begin{align}
				\E\bigl[\ell_\phi\bigl(X,f(Y)\bigr)\bigr]&=\E\bigl[\ell_\phi\bigl(X,\E[X|Y]\bigr)\bigr] \notag\\
				&+\E\bigl[\ell_\phi\bigl(\E[X|Y],f(Y)\bigr)\bigr].
				\label{thm:PythegorianIdentity}
			\end{align}%
		\item \emph{(Conditional Expectation is a Unique Bayesian Minimizer)} If $\E[X]<\infty$ and $\E[\phi(X)]<\infty$, then
			\begin{align}
				R_\phi(P_X, P_{Y|X}) &=\inf_{f:f\textup{ is measurable}}\E\bigl[\ell_\phi\bigl(X,f(Y)\bigr)\bigr]\notag \\
				&=\E\bigl[\ell_\phi\bigl(X,\E[X|Y]\bigr)\bigr].
				\label{eq:OptimalityOfCOnditionalExpectation}
			\end{align}%
		The optimizer in \eqref{eq:OptimalityOfCOnditionalExpectation} is unique almost surely~$P_Y$. 
	%\item \emph{(Coupling between Conditional Expectation and  Brebman Divergence)}   If   $F :\mathbb{R}^n \times  \mathbb{R}^n  \mapsto \mathbb{R}$ be a non-negative function such that $F(x,x)=0$ and assume that the partial derivatives $F_{x_i,x_j}$ are all continuous.   Suppose that for all random variables $X \in \mathbb{R}^n$,  if 
	%\begin{align}
	%\inf_{u \in \mathbb{R}^n}\E \left[  F(X,u)\right]=   \E \left[ F(X,\E[X])\right],
	%\end{align}
	%and $\E[X]$ is a unique minimizer.     Then,   $F(u,v)=  \ell_\phi(u,v)$ for some strictly convex and differentiable function $\phi: \mathbb{R}^n \mapsto \mathbb{R}$. 
	\end{enumerate}% 
\end{theorem}%
%
%{\color{red}[MG] We should provide a reference where these properties have been proven.}
%
%
%
%%%%%%%%%%%%%%%%%%%%%%%%%%%%%%%%%%%%%%%%%%%%%%%%%%%%%%%%%%%%%%%%%%%%%%%%%%%%%%%%%%%%%%%%%%%%%%%%%%%%%%%%%%%%%%%%%%%
\subsection{Moment Set} 
\begin{definition}\emph{(Moment Set)}
	Let $\bigl(\Omega,\sigma(\Omega)\bigr)$ be a measurable space and let $\mathcal{P}_{\mathsf{reg}}(\Omega)$ be the set of all regular probability measures over the sample space $\Omega$.\footnote{Recall that a probability measure is \emph{regular} if any element of the $\sigma$-algebra $\sigma(\Omega)$ can be approximated from below by compact measurable sets and from above by open measurable sets.} For any given $k\in\mathbb{N}$ fix measurable functions $f_1,\dots,f_k$ as well as real numbers $c_1,\dots,c_k$. Then, the set
	\begin{equation}%
		\mathcal{H}_k\coloneqq\bigl\{P\in\mathcal{P}_{\mathsf{reg}}(\Omega):\E_P[f_i(X)]\le c_i,\,1 \le i \le k\bigr\};
		\label{eq:DefinitionOfMomentSets}
	\end{equation}%
	that is, the set of regular probability measures with $k$ bounded moments, is called the \emph{moment set}. 
	\label{def:momentsset}
\end{definition}% 
\begin{remark}%
	The restriction to regular probability measures is rather mild. For example, Ulam's theorem \textup{\cite[Th.\, 7.1.4]{dudley2002real}} shows that a probability measure defined over a complete separable metric sample space (e.g., $\Omega=\mathbb{R}^n$) is regular.
\end{remark}%
%
%
%
%%%%%%%%%%%%%%%%%%%%%%%%%%%%%%%%%%%%%%%%%%%%%%%%%%%%%%%%%%%%%%%%%%%%%%%%%%%%%%%%%%%%%%%%%%%%%%%%%%%%%%%%%%%%%%%%%%%
\subsection{Conditional Expectation} 
Note that the conditional expectation $\E[X|Y]$ depends on the joint distribution $P_{XY}$ through the conditional $P_{X|Y}$. However, since in this paper $P_{Y|X}$ is fixed and $P_X$ varies, it is more convenient to treat $\E[X|Y]$ as a functional of $P_X$. Therefore, whenever we need to emphasize the dependence of the conditional expectation on the prior distribution, with a slight abuse of notation we will write $\E_{P_X}[X|Y]$. 
\begin{definition}\label{def:Tcomp} 
	Consider a fixed $P_{Y|X}$ and a set of probability distributions $\mathcal{F} \subseteq \mathcal{P}(\Omega)$. We say that Tweedie compatibility holds (with respect to $P_{Y|X}$ and $\mathcal{F}$), or T-compatibility for short,  if there exists an operator $f :\mathcal{Y}  \times \mathcal{P}(\Omega) \to \mathbb{R}^n$ such that for every $P_X \in \mathcal{F}$
	\begin{equation}
		\E_{P_X}[X|Y]= f(Y; P_Y)\text{ a.s.},
		\label{eq:Tweedy_Idenity} 
	\end{equation}%
	where $P_Y$ is the marginal distribution of $Y$ induced by $P_X$ (i.e., $\forall\mathcal{A}\in\sigma(\Omega):P_Y(\mathcal{A})= \E_{P_X}[ P_{Y|X}(\mathcal{A}|X)]$). 
\end{definition}%
% 
%{\color{red}[MG] $f:\mathcal{Y}\times\mathcal{P}(\mathcal{P})\to\mathbb{R}^n$ or $f:\mathcal{Y}\times\mathcal{P}(\Omega)\to\mathbb{R}^n$?}
T-compatibility simply says that the conditional expectation depends only on the marginal $P_Y$. An identity as in \eqref{eq:Tweedy_Idenity} is commonly known as Tweedie's formula \cite{robbins1956empirical,good1953population}. A family of distributions that are T-compatible is the following.
\begin{example} 
	Consider an exponential family 
	\begin{align}
	P_{Y|\Theta}(y|\theta) = h(y) \eu^{ \langle y, \theta \rangle -\psi(\theta) }
	\end{align}
	where $\theta$ is the natural parameter of the family, $h(y)$ the base measure, and $\psi(\theta)$ the log-partition function \textup{\cite{barndorff1980exponential}}. Now, if we set $X=\eu^{\Theta}$, then for $y\in\mathcal{Y}$
	\begin{equation}
		\E[X|Y=y]=\E[\eu^{\Theta}|Y=y]=\frac{h(y)}{h(y+1)}\frac{P_Y(y+1)}{P_{Y}(y)}.
	\end{equation}% 
	For a concrete example consider $\Omega=[0,\infty)$, $\mathcal{Y}=\mathbb{N}\cup\{0\}$, and let $P_{Y|X}(y|x)$ be Poisson transition probabilities; that is,
	\begin{equation}
	P_{Y|X}(y|x)=\frac{1}{y!} x^y \eu^{-x}\;,\;y\in \mathcal{Y}\;,\;x \in \Omega, 
	\end{equation}%
	where $x$ is the mean parameter. It is not difficult to check that $\Theta=\log(X)$ and $h(y)=\frac{1}{y!}$ so that  
	\begin{equation}
	\E[X|Y=y]= \frac{(y+1) P_{Y}(y+1)}{ P_{Y}(y)}\;,\;y \in \mathcal{Y}. 
	\end{equation}% 
\end{example}% 
%{\color{red}[MG] What is $\eta$ in (14)? Should be $\theta$? I would suggest we use a different letter for the log-partition function, as we were using $\phi$ for Bregman divergence. How about $\psi$?}
%
%
%
%%%%%%%%%%%%%%%%%%%%%%%%%%%%%%%%%%%%%%%%%%%%%%%%%%%%%%%%%%%%%%%%%%%%%%%%%%%%%%%%%%%%%%%%%%%%%%%%%%%%%%%%%%%%%%%%%%%%%%% 
\subsection{Problem Statement} 
We begin by listing assumptions that we are going to make throughout the rest of this paper.
\begin{assumption}\label{as:T1}%
	\begin{enumerate}[leftmargin=25pt]
		\item[]
		\item[(i)] $\mathcal{H}_k$ as defined in \eqref{eq:DefinitionOfMomentSets} is compact;
		\item[(ii)]  $R_\phi( P_X, P_{Y|X})$ is upper semicontinuous over $\mathcal{H}_k$;
		\item[(iii)] $|\mathcal{Y}|\le N$, $N\in\mathbb{N}$ (i.e., the support of the noisy observation $Y$ is finite);
		\item[(iv)] For every $y\in\mathcal{Y}$, $P_{Y|X}(y|x)$ is continuous in $x$ on the interior of $\Omega\subseteq\mathbb{R}^n$.	
	\end{enumerate}%
\end{assumption}%

Note that assumptions \textit{(i)}, \textit{(ii)} and \textit{(iv)} are not very restrictive. For example, \textit{(i)} and \textit{(ii)}  just guarantee the existence of a least favorable prior. Condition \textit{(iii)}, however, which imposes a restriction on the cardinality of the support of noisy observation $Y$, represents the main restriction in this work. In many situations, $\mathcal{Y}$ is indeed finite and in such case condition \textit{(iii)} is not a limitation. See Section~\ref{sec:binomial_ex} for an example.
 
It is also important to emphasize that we do not impose any conditions on the positions taken by the support $\mathcal{Y}$ of $Y$. Moreover, $N$ can be taken as large as needed. Thus, $Y$ can serve as an $N$-level quantization of some random vector $U$ fully supported on $\mathbb{R}$. See Section~\ref{sec:quantized_ex} for an example.
 
Another assumption that we may or may not make is the following.
\begin{assumption}\label{as:T2}
	T-compatibility, as defined in Definition~\ref{def:Tcomp}, holds with respect to $P_{Y|X}$ and $\mathcal{H}_k$. 
\end{assumption}%

Now, the objective of this work is to study the optimization problem 
\begin{equation}
	\sup_{P_X \in  \mathcal{H}_k}R_\phi(P_X, P_{Y|X}),
	\label{eq:problem}
\end{equation}% 
s.t. Assumption~\ref{as:T1} and potentially also Assumption~\ref{as:T2}. 
%
%
%
%%%%%%%%%%%%%%%%%%%%%%%%%%%%%%%%%%%%%%%%%%%%%%%%%%%%%%%%%%%%%%%%%%%%%%%%%%%%%%%%%%%%%%%%%%%%%%%%%%%%%%%%%%%%%%%%%%%%%%% 
\section{MAIN RESULT} 
\label{sec:Main_Result}
The main result of this work is the following. 
\begin{theorem}\emph{({Least Favorable Distribution})}\label{thm:BoundedOutPutChannel} 
	Let $N\in\mathbb{N}$ be finite, $k,n\in\mathbb{N}$ arbitrary but fixed, and \textup{Assumption~\ref{as:T1}} be fulfilled. Then, there exists a distribution $P^{\star}_{X}\in\mathcal{H}_k$ with the following properties: 
	\begin{itemize}[leftmargin=20pt]
		\item $\displaystyle{\max_{P_X\in \mathcal{H}_k }R_\phi( P_X, P_{Y|X})=R_\phi( P_X^\star, P_{Y|X})}$;  %\label{eq:Maximizing_stuff}
		\item  $P^{\star}_{X}\in\mathcal{H}_k$ is discrete with at most $N(k+1) (n+1)$ mass points (possibly containing mass points with individual coordinates equal to $\pm\infty$);
		\item $P^{\star}_{X}\in\mathcal{H}_k$ is discrete with at most $N(k+1)$ mass points if in addition \textup{Assumption~\ref{as:T2}} is fulfilled.
	\end{itemize}% 
\end{theorem}%

Theorem~\ref{thm:BoundedOutPutChannel} allows us to move the optimization in \eqref{eq:Main_Optimization} from the space of probability distributions to $\mathbb{R}^{n d+d}$, where $d\le N(k+1)(n+1)$ or $d\le N(k+1)$ in case Assumption~\ref{as:T2} is fulfilled. More specifically, we can parameterize the input distribution by a vector containing the sought after probability masses together with their locations:
\begin{align}
	P_X\,\Rightarrow\,\mathbf{x}&=[\mathbf{x}_1,\mathbf{x}_2]^\transpose\in\mathbb{R}^{nd+d} \notag \\ 
	&=\Big[\hspace{-8pt}\underbrace{x_1^\transpose,x_2^\transpose,\dots,x_d^\transpose}_{\substack{\text{points of the support}\\ x_i\in \mathbb{R}^n, 1 \le i \le d}},\,\underbrace{p_1, p_2,\dots, p_d}_{\text{probability masses}}\hspace{-3pt}\Big]^\transpose.
	\label{eq:parametrizing_probabilities} 
\end{align}%
Working in $\mathbb{R}^{nd+d}$ has the huge advantage that in order to find a least favorable distribution we can employ numerical methods such as projected gradient ascent \cite{shalev2014understanding}. More details on this will be given in Section~\ref{sec:Simulations}. 

Perhaps somewhat remarkable is that under Assumption~\ref{as:T2} we obtain a bound on the cardinality of the support of a least favorable distribution that is \emph{independent} of the dimension of $\Omega$. This might have potential applications to the case where   $\Omega$ is an infinite dimensional set.  

We next show that under Assumption~\ref{as:T1} and some extra conditions the dimensionality can further be reduced. The corresponding proof can be found in Section~\ref{sec:proofprop1} of the supplementary material. 
\begin{prop} \label{prop:Further_Reduction}
	 Let $N\in\mathbb{N}$ be finite, $k,n\in\mathbb{N}$ arbitrary but fixed, and \textup{Assumption~\ref{as:T1}} be fulfilled. Then, $P^{\star}_{X}$ is discrete with at most $(n+1)(N-1)+k+1$ mass points if
\begin{itemize}[leftmargin=20pt]
	\item $\Omega\subset\mathbb{R}^n$ is compact and $f_1,\dots,f_k$ are bounded and continuous on $\Omega$; or %, then $P^{\star}_{X}$ has at most $N+k$ finite mass points; 
	\item $f_1,\dots,f_k$ are  continuous on $\Omega=\mathbb{R}^n$ and are such that for every $P_X$ with a finite number of mass points $\E_{P_X}[f_i(X)]<\infty$ implies $P_{X_j}(+\infty)=P_{X_j}(-\infty)=0$ for all $1\leq j\leq n$,  with $X_j$ the $j$-th coordinate of $X$. % then $P^{\star}_{X}$ has at most $N+k$ finite mass points.
	\end{itemize}% 
\end{prop}% 
\begin{remark} 
	If functions $f_1,\dots,f_k$ in the definition of moment set $\mathcal{H}_k$ prevent the occurrence of mass points at $\pm \infty$, then the bound on the number of points can be reduced to $(n+1)(N-1)+k+1$. An example of such a function is $f:\mathbb{R}\to\mathbb{R}$, $f(x)=|x|^r$, $r>0$, which naturally forces probability measures with a finite number of mass points to have mass points at $\pm \infty$ with zero probability.  
\end{remark}% 

Finally, we note that Theorem~\ref{thm:BoundedOutPutChannel} does not guarantee that the least favorable distribution is unique or that every least favorable distribution is discrete. It only guarantees that there exists a least favorable distribution that is discrete.
%
%
%
%%%%%%%%%%%%%%%%%%%%%%%%%%%%%%%%%%%%%%%%%%%%%%%%%%%%%%%%%%%%%%%%%%%%%%%%%%%%%%%%%%%%%%%%%%%%%%%%%%%%%%%%%%%%%%%%%%%%%%% 
\section{PROOF OF THE MAIN THEOREM} 
\label{sec:Proof_main}
%
%
%
%%%%%%%%%%%%%%%%%%%%%%%%%%%%%%%%%%%%%%%%%%%%%%%%%%%%%%%%%%%%%%%%%%%%%%%%%%%%%%%%%%%%%%%%%%%%%%%%%%%%%%%%%%%%%%%%%%%%%%%%%%%
\subsection{Preliminaries}
Before actually proving Theorem~\ref{thm:BoundedOutPutChannel}, we provide some preliminary definitions and results that will help us to accomplish this. 
%
%
%
%%%%%%%%%%%%%%%%%%%%%%%%%%%%%%%%%%%%%%%%%%%%%%%%%%%%%%%%%%%%%%%%%%%%%%%%%%%%%%%%%%%%%%%%%%%%%%%%%%%%%%%%%%%%%%%%%%%%%%%%%%%
\subsubsection{Weak Convergence and Weak Continuity}
It is well known that there exist several definitions of the convergence of a sequence of probability measures. One is weak convergence, which provides a given space of probability measures with a topology.
\begin{definition}
	A sequence of probability measures $\{P_n\}_{n\in\mathbb{N}}$ is said to \emph{converge weakly} to probability measure $P$ if for every bounded and continuous function $\psi$  
	\begin{align}
		\lim_{n \to \infty}\E_{P_n}\bigl[\psi(X)\bigr]\to\E_{P}\bigl[\psi(X)\bigr].
	\end{align}%
\end{definition}%

Another main ingredient of our considerations are linear functionals. The following theorem gives a necessary and sufficient condition for a linear functional to be weakly continuous \cite[Lemma 2.1]{huber1981robust}. 
\begin{theorem} \emph{(Weak Continuity of Linear Functionals)} \label{thm:ContinuityOfLinearFunctionals} A linear functional $L:\mathcal{P}{(\Omega)}\to\mathbb{R}$ is \emph{weakly continuous} on $\mathcal{P}{(\Omega)}$ if and only if there exists a bounded and continuous function $\psi$ such that $L$ can be represented as
	\begin{equation*}
		L(P)= \E_P\bigl[\psi(X)\bigr].
	\end{equation*}%
\end{theorem}%
%{\color{red}[MG] What does "some" precisely mean in the last part of the sentence? Does it mean there "exists" a bounded and continuous function $\phi$ such that... or does it mean "for all" bounded and continuous functions?}  {\color{blue} [AD] EXISTS} 
%
%
%
%%%%%%%%%%%%%%%%%%%%%%%%%%%%%%%%%%%%%%%%%%%%%%%%%%%%%%%%%%%%%%%%%%%%%%%%%%%%%%%%%%%%%%%%%%%%%%%%%%%%%%%%%%%%%%%%%%%%%%%%%%%
\subsubsection{Linear Programming} 
The extreme value theorem for real-valued continuous functions over compact intervals is one of the most celebrated results of calculus. The following theorem is a generalization to compact topological spaces \cite[Sec.\,2.13]{luenberger1997optimization}.
\begin{theorem}\emph{(Extreme Value Theorem)}\label{thm:ExtremValueTheorem} For every compact topological space $\mathcal{P}$ and every upper semicontinuous (lower semicontinuous) functional $f:\mathcal{P}\to\mathbb{R}$ 
	\begin{equation*}
		\sup_{P\in\mathcal{P}}f(P)=\max_{P\in\mathcal{P}}f(P)\quad\left(\inf_{P\in\mathcal{P}}f(P)=\min_{P\in\mathcal{P}}f(P)\right). 
	\end{equation*}
Moreover, if $f$ is strictly concave (strictly convex) the maximizer (minimizer) is unique. 
\end{theorem}%
\begin{definition}  
	An \emph{extreme point} of any convex set $\mathcal{S}$ is a point $x\in\mathcal{S}$ that cannot be represented as $x = (1-\alpha)y+\alpha z$ with $y,z\in \mathcal{S}$ and $\alpha\in(0,1)$. We denote the set of all extreme points of $\mathcal{S}$ as $\ex(\mathcal{S})$.
\end{definition}%

The following result states that when maximizing a linear functional over a moment set it is sufficient to focus on its extreme points \cite[Th.\,3.2]{winkler1988extreme}.
\begin{theorem} \emph{(Linear Programming)} \label{thm:LinearProgramming} Let $L:\mathcal{P}{(\Omega)}\to\mathbb{R}$ be a linear functional. Then,
	\begin{equation*}
		\sup_{P \in \mathcal{H}_k} L(P)= \sup_{P\in\ex(\mathcal{H}_k)} L(P).
	\end{equation*}% 
\end{theorem}%
Note that Theorem~\ref{thm:LinearProgramming} only requires $L$ to be linear and not necessarily continuous.
%
%
%
%%%%%%%%%%%%%%%%%%%%%%%%%%%%%%%%%%%%%%%%%%%%%%%%%%%%%%%%%%%%%%%%%%%%%%%%%%%%%%%%%%%%%%%%%%%%%%%%%%%%%%%%%%%%%%%%%%%%%%%%%%%
\subsubsection{Extreme Points of a Moment Set} 
For proving the main result of this paper, the following theorem will be of particular importance \cite[Th.\,2.1]{winkler1988extreme}. 
\begin{theorem}\emph{(Extreme Points of a Moment Set)}\label{thm:ExtremPointsOfMomentSets} 
	For given $k\in\mathbb{N}$ let moment set $\mathcal{H}_k$ be defined as in Definition~\ref{def:momentsset}. Then, the following holds: 
	\begin{itemize}
		\item  $\mathcal{H}_k$ is convex and the extreme points of $\mathcal{H}_k$ are
			\begin{equation}%
				\ex(\mathcal{H}_k)\subseteq\overline{\ex(\mathcal{H}_k)},
				\label{eq:BoundOnTheSefOfExtremPoints}
			\end{equation}%
			where
			\begin{align*}
			  &\overline{\ex(\mathcal{H}_k)} \notag\\
			  &\coloneqq\biggl\{\! P\in \mathcal{H}_k:P= \sum_{i=1}^m \alpha_i \delta_{x_i}, x_i\in\Omega, \alpha_i\in[0,1],  \notag\\
			  &\hphantom{aaaa}\sum_{i=1}^m \alpha_i=1\,,\,1\le m\le k+1\;\,\wedge\;\,\textup{the vectors}\notag\\[2pt]
			  &\hphantom{aaaa,}[f_1(x_i),\dots,f_k(x_i),1]^\transpose\,,\,1\leq i\leq m,\, \textup{are}\notag\\[2pt]
			  &\hphantom{aaaa,}\textup{linearly independent}\biggr\};
			\end{align*}%
		\item If the moment conditions in \eqref{eq:DefinitionOfMomentSets} are fulfilled with equality, then \eqref{eq:BoundOnTheSefOfExtremPoints} holds with equality. 
	\end{itemize}% 
\end{theorem}%

\begin{remark}%
	Theorem~\ref{thm:ExtremPointsOfMomentSets} is also valid in case we have a constraint on the support but no moment constraint (i.e., $k=0$). For example, let $\Omega=\mathcal{B}_0(r)$ for some $r>0$. Then,
	\begin{equation*}%
		\ex(\mathcal{H}_0)=\bigl\{P\in\mathcal{H}_0: P=\delta_{x}, x\in\mathcal{B}_0(r)\bigr\}. 
	\end{equation*}% 
	In case of a second moment constraint (i.e., $\mathcal{H}_1=\{P\in\mathcal{P}_{\mathsf{reg}}(\Omega):\E_P[\|X\|^2]\le c\}$, $c\in\mathbb{R}$) we have
	\begin{align*}
		&\ex(\mathcal{H}_1)\subseteq\overline{\ex(\mathcal{H}_1)}\\
		&=\bigl\{P\in\mathcal{H}_1: P= (1-\alpha) \delta_{x_1}+\alpha \delta_{x_2},\alpha \in [0,1], \notag\\
		&  \qquad x_1,x_2\in\Omega, \|x_1\| \neq \|x_2\|\bigr\}. 
	\end{align*}%
\end{remark}%
%
%{\color{red}[MG] Note that I've replaced radius $R$ in $\mathcal{B}_0(R)$ by $r$ (i.e., $\mathcal{B}_0(r)$ so that it gets not confused with the Bayesian risk. I also added the definition of a closed ball to the notations section.}
%
%
%
%%%%%%%%%%%%%%%%%%%%%%%%%%%%%%%%%%%%%%%%%%%%%%%%%%%%%%%%%%%%%%%%%%%%%%%%%%%%%%%%%%%%%%%%%%%%%%%%%%%%%%%%%%%%%%%%%%%%%%%%%%%
\subsubsection{Dubins' Theorem} 
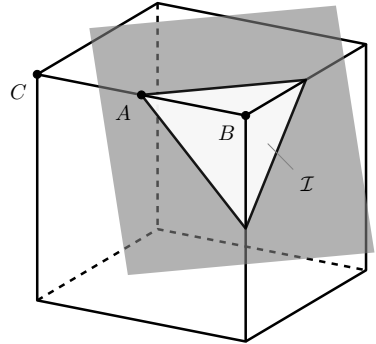
\begin{figure}[!t]
	\centering
	\scalebox{0.8}{\input{FIG/cube_plane.pgf}}
	\caption{Example of Dubins' Theorem for $n=1$. Cube $\mathcal{K}\subset\mathbb{R}^3$, which is a linearly closed and linearly bounded convex set, intersects with a hyperplane in a triangle $\mathcal{I}$. An extreme point/vertex of $\mathcal{I}$ (e.g., point $A$) belongs to an edge of $\mathcal{K}$. Thus, $A$ is the convex combination of $n+1=2$ extreme points of $\mathcal{K}$; that is, a convex combination of points $B$ and $C$.}%
	\label{fig:dubins}%
\end{figure}

\begin{definition}%
	A convex set $\mathcal{S}$ of a vector space $\mathcal{V}$ is called \emph{linearly closed (linearly bounded)} if every straight line
	intersects with $\mathcal{S}$ on a closed (bounded) subset of that line. %for any two points $x,y\in\mathcal{S}$, $x\neq y$, the intersection of $\mathcal{S}$ with the line through $x$ and $y$ has two extreme points.
\end{definition}%
With this definition in hand, a powerful theorem proven by Dubins \cite{dubins1962extreme} is the following.
\begin{theorem}\label{thm:DubinsTheorem} \emph{(Dubins' Theorem)} Let $f:\mathcal{V}\to\mathbb{R}$ be a linear functional over a vector space $\mathcal{V}$ and let
	\begin{equation*}%
		\mathcal{L}=\{v\in\mathcal{V} : f(v)=c\},
	\end{equation*}%
	for some $c\in\mathbb{R}$, be a hyperplane formed by $f$. Moreover, let $\mathcal{I}$ be the intersection of a linearly closed and linearly
	bounded convex set $\mathcal{K}\subset\mathcal{V}$ with $n$ hyperplanes. Then, every extreme point of $\mathcal{I}$ is a
	convex combination of at most $n + 1$ extreme points of $\mathcal{K}$. 
\end{theorem}% 
A remarkable property of Theorem~\ref{thm:DubinsTheorem} is that it also holds for the infinite dimensional case.  See Fig.~\ref{fig:dubins} for a finite-dimensional example of Dubins' Theorem.
%
%
%
%%%%%%%%%%%%%%%%%%%%%%%%%%%%%%%%%%%%%%%%%%%%%%%%%%%%%%%%%%%%%%%%%%%%%%%%%%%%%%%%%%%%%%%%%%%%%%%%%%%%%%%%%%%%%%%%%%%%%%%%%%%
\subsection{Proof of Theorem~\ref{thm:BoundedOutPutChannel}: Existence of a Solution}
Recall that the first part of the main result states that with Assumption~\ref{as:T1} a least favorable prior exists. Observe that this follows from Theorem~\ref{thm:ExtremValueTheorem} due to the semicontinuity of Bayesian risk $R_\phi(P_X,P_{Y|X})$ and the compactness of $\mathcal{H}_k$. Hence, the supremum in (\ref{eq:problem}) is attained by some $P_{X}^\star\in\mathcal{H}_k$.   
%
%
%
%%%%%%%%%%%%%%%%%%%%%%%%%%%%%%%%%%%%%%%%%%%%%%%%%%%%%%%%%%%%%%%%%%%%%%%%%%%%%%%%%%%%%%%%%%%%%%%%%%%%%%%%%%%%%%%%%%%%%%%%%%%
\subsection{Proof of Theorem~\ref{thm:BoundedOutPutChannel}: Assumptions~\ref{as:T1} \& \ref{as:T2}} \label{sec:Proof_Under_T2} 
We first proof the second part of Theorem~\ref{thm:BoundedOutPutChannel} where Assumptions~\ref{as:T1} and \ref{as:T2} are both fulfilled as it is easier. 

As a least favorable prior, $P_X^\star$, exists (not necessarily unique), let $P_Y(\cdot ;\star)$ and $\E_{\star}[X|Y]:\mathcal{Y}\to\Omega$ denote the marginal of $Y$ and the conditional expectation of $X$ given $Y$ induced by $P_X^\star$.\footnote{More precisely, for given $P_X^\star$, $P_Y(\cdot;\star)$ and $\mathbb{E}_\star[X|Y]$ are just shorthand for $P_Y(\cdot;P_X^\star)$ and $\mathbb{E}_{P_X^\star}[X|Y]$.} Moreover, we define
\begin{align}
	\mathcal{P}^\star &\coloneqq \bigl\{P_X\in\mathcal{H}_k: P_Y(y;\star)=P_Y(y;P_X),\forall y\in\mathcal{Y}\bigr\}\notag\\
	&\hphantom{:}=\bigl\{P_X \in \mathcal{H}_k: c_y^\star= P_Y(y ; P_X) , \forall y \in \mathcal{Y}\bigr\}
\end{align}% 
as the set of least favorable priors that induce $P_Y(y ; \star)$, where for ease of notation $c_y^\star\coloneqq P_Y(y;\star)$.  

Next, note that as a consequence of Assumption~\ref{as:T2}, if $P_X, Q_X \in  \mathcal{P}^\star $, then for every $y\in\mathcal{Y}$
\begin{equation}
\E_{P_X}[X|Y=y]=\E_{Q_X }[X|Y=y]=\E_{\star}[X|Y=y].
\end{equation}%
Furthermore, observe that $\mathcal{P}^\star$ is the intersection of $\mathcal{H}_k$ with $N-1$ hyperplanes of the form
\begin{equation}
\mathcal{L}_i=\bigl\{P_X:c_{y_i} =\E_{P_X}[P_{Y|X}(y_i|X)]\bigr\}, 1\leq i\leq N-1,
\label{eq:hyperplanceFromTheOutput_T2}
\end{equation}%
where we have used that $P_Y(y ; P_X)= \E_{P_X}\bigl[ P_{Y|X}(y|X)\bigr]$, $y\in\mathcal{Y}$.
Note that we omitted hyperplane $\mathcal{L}_N$ in \eqref{eq:hyperplanceFromTheOutput_T2} as in the space of probability distributions everything sums up to one so that $\mathcal{L}_N$ is redundant. Note also that each $\mathcal{L}_i$ is a closed set, which follows from Theorem~\ref{thm:ContinuityOfLinearFunctionals} and the fact that $P_{Y|X}( y_i| x)$ is bounded and continuous in $x$ for each $y_i$ (i.e., sets defined by continuous functions are closed).

Next, observe that
\begin{align}
\max_{P_X\in\mathcal{H}_k }\E_{P_{X}}&\bigl[\ell_{\phi}\bigl(X,\E_{P_X}[X|Y]\bigr)\bigr]\notag\\
&\stackrel{a)}{=}\max_{P_X \in \mathcal{P}^\star}\E_{P_{X}}\bigl[\ell_{\phi}\bigl(X,\E_{P_X}[X|Y]\bigr)\bigr]\\
&\stackrel{b)}{=}\max_{P_X \in \mathcal{P}^\star}\E_{P_{X}}\bigl[\ell_{\phi}\bigl(X,\E_{\star}[X|Y]\bigr)\bigr]\\
&\stackrel{c)}{=}\max_{P_X \in\mathrm{ex}(\mathcal{P}^\star)}\E_{P_{X}}\bigl[\ell_{\phi}\bigl(X,\E_{\star}[X|Y]\bigr)\bigr], 
\end{align}%
where $a)$ follows from using the fact that $\mathcal{P}^\star$ contains a least favorable prior, $b)$ from the fact that  $\E_{P_X}[X|Y]=\E_{\star}[X|Y]$ for every $P_X\in\mathcal{P}^\star$, and $c)$ from Theorem~\ref{thm:LinearProgramming} by observing that $P_X \mapsto\E_{P_{X}}[\ell_{\phi}(X,\E_{\star}[X|Y])]$ is linear over $\mathcal{P}^\star$ and therefore $\E_{\star}[X|Y]$ does not change on $\mathcal{P}^\star$.

Finally, recall that $\mathcal{P}^\star$ consists of the intersection of $N-1$ hyperplanes, defined in \eqref{eq:hyperplanceFromTheOutput_T2}, with $\mathcal{H}_k$. Thus, as $\mathcal{P}^\star$ is a subset of $\mathcal{H}_k$, it follows from Theorem~\ref{thm:DubinsTheorem} that every extreme point of $\mathcal{P}^\star$ (or every point of $\mathrm{ex}(\mathcal{P}^\star)$) can be represented by a convex combination of at most $(N-1)+1=N$ extreme points of $\mathcal{H}_k$. Due to Theorem~\ref{thm:ExtremPointsOfMomentSets}, however, the extreme points are discrete distributions with at most $k+1$ mass points, so that $P_X^\star$ consists of at most $N(k+1)$ mass points.   
%
%
%
%%%%%%%%%%%%%%%%%%%%%%%%%%%%%%%%%%%%%%%%%%%%%%%%%%%%%%%%%%%%%%%%%%%%%%%%%%%%%%%%%%%%%%%%%%%%%%%%%%%%%%%%%%%%%%%%%%%%%%%%%%%
\subsection{Proof of Theorem~\ref{thm:BoundedOutPutChannel}: Assumption~\ref{as:T1} only} 
The key to the proof of Theorem~\ref{thm:BoundedOutPutChannel} under Assumptions~\ref{as:T1} and \ref{as:T2} is the construction of a set $\mathcal{P}^\star$ over which the conditional expectation does not change and which can be written as an intersection of finitely many hyperplanes with a moment set. Assumption~\ref{as:T2} allows us to do this without the dependance on the dimension $n$ of $\Omega$. In the general case (i.e., without Assumption~\ref{as:T2}), however, this does not seem to be possible, which is why we have to construct $\mathcal{P}^\star$ differently. 

Towards this end, let $\E_{\star}[X|Y]:\mathcal{Y}\to\Omega$ as in the previous subsection be induced by some least favorable prior $P_X^\star$. Furthermore, note that the $j$'s component of $\E_{\star}[X|Y]$ can be written as
\begin{align}
	\E_{\star}[X_j|Y=y]&= \frac{\E_{\star}\bigl[X_j P_{Y|X_j}(y|X_j)\bigr]}{P_{Y}(y;\star)}\\
	&=\frac{\E_{\star}\bigl[X_j P_{Y|X_j}(y|X_j)\bigr]}{\E_{\star}\bigl[P_{Y|X}(y|X)\bigr]}
	\label{eq:Ration_representation}
\end{align}%
for every $y\in\mathcal{Y}$ and $j=1,\dots,n$. The expression in \eqref{eq:Ration_representation} implies that the conditional expectation is a ratio of two linear functionals. Therefore, the set $\mathcal{P}^\star$ can be constructed as follows:
\begin{align}
	\mathcal{P}^\star=\bigl\{&P_X\in\mathcal{H}_k:c_{ji}^\star=\E_{P_X}\bigl[X_j P_{Y|X_j}(y_i|X_j)\bigr]\,\,\land\notag\\
	&c_i^\star=P_{Y}(y_i; P_X) \,   1 \le j \le n, \, 1 \le i \le N\bigr\},
	\label{eq:ConstraintSetfjalkfj;aldjflalkdfal}
\end{align}%
where for ease of notation
\begin{align}
	c_i^\star &\coloneqq P_{Y}(y_i;\star), \, 1 \le i \le N-1, \label{eq:Omitting_One_plane} \\
	c_{ji}^\star &\coloneqq\E_{\star}\bigl[X_jP_{Y|X_j}(y_i|X_j)\bigr],\,1 \le j \le n, \, 1 \le i \le N. 
\end{align}%
Again, we omitted the hyperplane $c_N^\star$ in \eqref{eq:Omitting_One_plane} due to the same reasons as mentioned in the context of (\ref{eq:hyperplanceFromTheOutput_T2}). 

By construction $\E_{\star}[X|Y=y_i]=\E_{P_X}[X|Y=y_i]$ for every $P_X\in\mathcal{P}^\star$. In addition, the set $\mathcal{P}^\star$ is the intersection of the $(n+1)N-1$ hyperplanes
\begin{equation}
	\mathcal{L}^{(a)}_i= \bigl\{P_X:c_i^\star=\E_{P_X}\bigl[P_{Y|X}(y_i|X)\bigr]\bigr\}
	\label{eq:hyperplanceFromTheOutput}
\end{equation}%
for $1\le i\le N-1$ and
\begin{equation}
	\mathcal{L}^{(b)}_{ij}=\bigl\{P_X:c_{ji}^\star=\E_{P_X}\bigl[ X_j P_{Y|X_j}(y_i|X_j)\bigr]\bigr\}
	\label{eq:hyperplanceNumeratorOfConditionalExpectation}
\end{equation}%
for $1 \le i\le N$ and $1 \le j \le n$, respectively. Now, at this point, following the same line of arguments as in Section~\ref{sec:Proof_Under_T2} we arrive at the conclusion that $P_X^\star$ consists of at most $(k+1) (n+1)N$ mass points.
%
%
%
%
%%%%%%%%%%%%%%%%%%%%%%%%%%%%%%%%%%%%%%%%%%%%%%%%%%%%%%%%%%%%%%%%%%%%%%%%%%%%%%%%%%%%%%%%%%%%%%%%%%%%%%%%%%%%%%%%%%%%%%%%%%%
\section{NUMERICAL EXAMPLES}\label{sec:Simulations} 
To demonstrate the findings of this paper, in this section we present two numerical examples carried out with the projected gradient ascent method. In the following subsection, we first provide a description of that method where for the sake of simplicity we only focus on the case of a support constraint. That is, $\Omega$ is bounded and moment constraints are not present (i.e., $k=0$). As a result, the first bound on the cardinality becomes $N(n+1)$ and the second bound just $N$. 
%
%
%
%%%%%%%%%%%%%%%%%%%%%%%%%%%%%%%%%%%%%%%%%%%%%%%%%%%%%%%%%%%%%%%%%%%%%%%%%%%%%%%%%%%%%%%%%%%%%%%%%%%%%%%%%%%%%%%%%%%%%%%%%%%
\subsection{Projected Gradient Ascent} 
Firm upper bounds on the number of mass points, such as the one given in Theorem~\ref{thm:BoundedOutPutChannel}, allow us to carry out the optimization over the space $\mathbb{R}^{nd+d}$, where $d$ denotes the number of mass points, instead over the space of probability distributions. As mentioned in the discussion after Theorem~\ref{thm:BoundedOutPutChannel}, working in $\mathbb{R}^{nd+d}$ has the advantage that we can employ numerical methods such as projected gradient ascent \cite{shalev2014understanding}. A quick sketch of how to use projected gradient ascent for finding a least favorable prior follows next.

As described in \eqref{eq:parametrizing_probabilities}, we parameterize prior distribution $P_X$ by a vector $\mathbf{x}=[\mathbf{x}_1,\mathbf{x}_2]^\transpose\in\mathbb{R}^{nd+d}$ containing the sought after probability masses together with their locations. From Theorem~\ref{thm:BoundedOutPutChannel} we know that $d$ can be at most $N(n+1)$, or $N$ if in addition Assumption~\ref{as:T2} holds. Now, let
\begin{equation}
	g(\mathbf{x})\coloneqq R_\phi(P_X, P_{Y|X})
\end{equation}%
and define 
\begin{equation}
	\mathcal{G}\coloneqq\bigl\{\mathbf{x}\in\mathbb{R}^{nd+d}:\mathbf{x}_1\in \Omega\,,\,\mathbf{x}_2\in\mathcal{S}\bigr\}
\end{equation}%
as the constraint set with $\mathcal{S}$ denoting the probability simplex. 
In view of Theorem~\ref{thm:BoundedOutPutChannel}, we then have
\begin{equation}
	\max_{P_X: X \in \Omega} R_\phi(P_X, P_{Y|X}) = \max_{\mathbf{x}\in\mathcal{G}} g(\mathbf{x}). 
\end{equation}%    
Starting from an initial point $\mathbf{x}^{(1)}\in\mathcal{G}$, projected gradient ascent iterates the following equation until a stopping criterion is met: 
\begin{equation}
	\mathbf{x}^{(t+1)}=\mathsf{proj}_\mathcal{G}\bigl(\mathbf{x}^{(t)}+\lambda\nabla g(\mathbf{x}^{(t)})\bigr), \, t\in \mathbb{N}\,,
\end{equation} 
where $\lambda>0$ is some step size, $\nabla g({\mathbf x})$ the gradient of $g$, and $\mathsf{proj}_\mathcal{G}(\cdot)$ the projection operator that tries to find a point $\mathbf{x}^{(t+1)}\in\mathcal{G}$ that is closest to $\mathbf{x}^{(t)}+\lambda\nabla g(\mathbf{x}^{(t)})$ in squared Euclidean distance.

If $\Omega$ is convex, the projection of $\mathbf{x}_1$ onto $\Omega$ can be done efficiently by using the alternating projection method \cite{bauschke1996projection}, whereas an efficient implementation of the projection of $\mathbf{x}_2$ onto the probability simplex with complexity $\mathcal{O}(d \log d)$ can be found in \cite{wang2013projection}. 

In order to implement the projected gradient ascent algorithm described above, we obviously have to compute the gradient of $g$. The following result provides the gradient for the special case of $n=1$ and $\phi$ being the squared error loss. The corresponding proof can be found in Section~\ref{sec:Grad_MMSE} of the supplementary material. 
\begin{prop} 
	Let $\phi:\Omega\to\mathbb{R}$, $\phi(x)=x^2$. Then, if $g$ is differentiable we have for $i=1,\dots,N$
	\begin{align}
		\frac{\partial}{\partial p_i}g(\mathbf{x})&=\E\bigl[\bigl(x_i-\E[X|Y]\bigr)^2| X=x_i\bigr],  \\
		\frac{\partial}{\partial x_i}g(\mathbf{x})&=2 p_i\bigl(x_i-\E\bigl[\E[X|Y]\big|X=x_i\bigr]\bigr)\notag\\
		&+p_i\sum_{j=1}^NP_{Y|X}'(y_j|x_i)\Bigl(\bigl(\E[X| Y=y_j]\bigr)^2\notag\\
		&-2 x_i\E[X|Y=y_j]\Bigr). 
	\end{align}% 
\end{prop}% 
%
%
%
%%%%%%%%%%%%%%%%%%%%%%%%%%%%%%%%%%%%%%%%%%%%%%%%%%%%%%%%%%%%%%%%%%%%%%%%%%%%%%%%%%%%%%%%%%%%%%%%%%%%%%%%%%%%%%%%%%%%%%%%%%%
\subsection{Binomial Model}\label{sec:binomial_ex}
Consider conditional probability mass function 
\begin{equation}
	P_{Y|X}(y|x)= \binom{m}{y} x^y (1-x)^{m-y}, \,  x \in \Omega,\, y \in \mathcal{Y}. 
\end{equation}% 
Here, $x$ is the success probability, $\Omega=[0,1]$, and $\mathcal{Y}=\{0,1,\ldots, m\}$, $m\in\mathbb{N}$. We treat parameter $m$ of this distribution as known so that $N=m+1$. This binomial model is very popular and has a wide range of applications (see \cite{neter1988applied} for details).

We are now interested in estimating the success probability, which we model as a random variable $X\in\Omega$ with distribution $P_X$. To compute least favorable prior $P_X^\star$, we use the gradient ascent algorithm described in the previous subsection. 

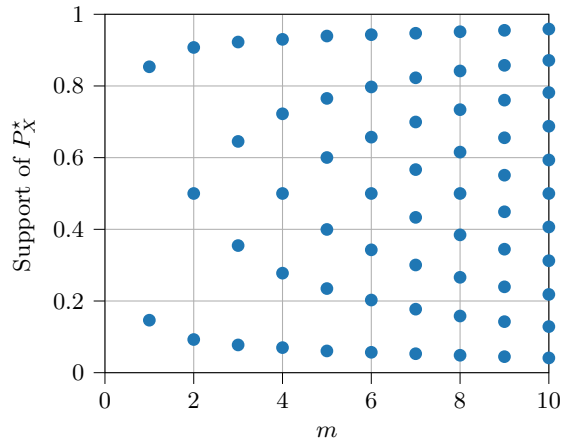
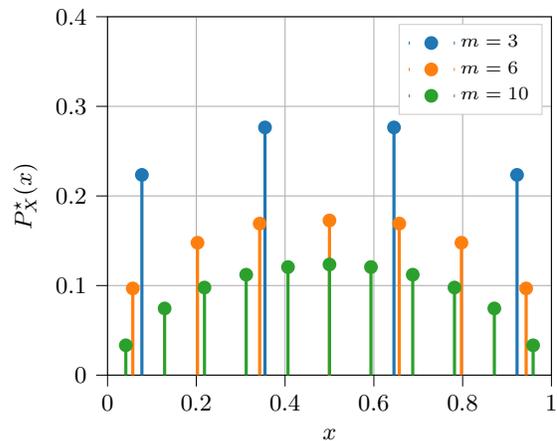
\begin{figure}[t!]
	\centering
	\begin{subfigure}[t]{.9\linewidth}
		\centering
		\input{FIG/PositionBinomial.pgf}\vspace{-8pt}
		\caption{Support of least favorable prior $P_X^\star$ for different values of $m$.}
		\label{fig:Position}
	\end{subfigure}%
	~
	\
	\

	\begin{subfigure}[t]{.9\linewidth}
		\centering
		\input{FIG/BinomialPMFm3610.pgf}\vspace{-8pt}
		\caption{Least favorable priors for $m=3,6,10$. }
		\label{fig:Prob}
	\end{subfigure}%
	\caption{Least favorable priors for $P_{Y|X}$ being binomial with parameter $m$.}
	\label{fig:Distr_Binomial}
\end{figure}

Fig.~\ref{fig:Distr_Binomial} depicts the outcome of the algorithm for $m$ ranging from 1 to 10. More specifically, Fig.~\ref{fig:Position} shows the support of $P_X^\star$ as a function of $m$. It is interesting to note that the support is uniformly spaced but $P_X^\star$ is not the uniform distribution, which can be seen from Fig~\ref{fig:Prob} for $m=3,6,10$.  
%
%
%
%%%%%%%%%%%%%%%%%%%%%%%%%%%%%%%%%%%%%%%%%%%%%%%%%%%%%%%%%%%%%%%%%%%%%%%%%%%%%%%%%%%%%%%%%%%%%%%%%%%%%%%%%%%%%%%%%%%%%%%%%%%
\subsection{Quantized Gaussian}\label{sec:quantized_ex}
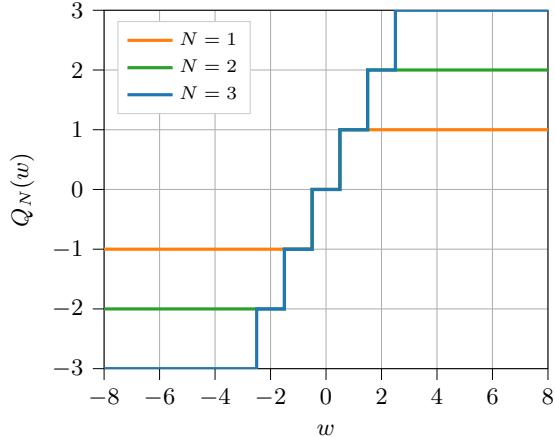
\begin{figure}[t!]
	\centering
	\input{FIG/QuantFunction.pgf}
	\caption{Quantization function $Q_N$ for $N=1,2,3$.}
	\label{eq:quantatizatio_Fucntion}
\end{figure}

As another example consider the scenario in which we seek to estimate a random variable $X$ that is embedded in Gaussian noise:
\begin{equation}
	W=X+Z\;\,,\;\,Z\sim\mathcal{N}(0,1).
\end{equation}%
Suppose that instead of observing $W$ directly, we only have access to a quantized version of $W$, which we denote as $Y$. Specifically, consider uniform quantization with clipping; that is, for a fixed integer $N$
\begin{equation}
	y=Q_N(w)=\begin{cases} \lfloor w  \rceil\,\quad &|w|<N\\ \mathrm{sign}(w) N &\text{else}\end{cases}, 
\end{equation}% 
where $\lfloor \cdot  \rceil$ rounds to the nearest integer. Examples of $Q_N$ are depicted in Fig.~\ref{eq:quantatizatio_Fucntion}.

Note that in this scenario, $Y$ is supported on $\mathcal{Y}=\{-N, \ldots, 0, \ldots, N\}$. Moreover, the conditional distribution of $Y$ given $X$ is for $N\geq 1$ given by
\begin{align}
	&P_{Y|X}(y|x) \notag\\
	&\hphantom{aaa}=\begin{cases}\Phi\parentheses*{-N -x +\frac{1}{2}}, &y=-N\\
 	\Phi\parentheses*{y-x+\frac{1}{2}}-\Phi\parentheses*{-x-\frac{1}{2}},  &|y|\leq N-1\\
 	\Phi\parentheses*{-N +x +\frac{1}{2}}, & y=N\end{cases},  
 	\label{Eq:Gaussian_Case}
\end{align}%
where $\Phi$ denotes the cumulative distribution function of a standard Gaussian. Using Theorem~\ref{thm:BoundedOutPutChannel}, we have that the cardinality of the support of $P_{X}^\star$ is bounded by $2N+1$.

\begin{figure}[t!]
	\begin{subfigure}[t]{.9\linewidth}
	\center
		\input{FIG/Gn4A5.pgf}\vspace{-8pt}
		\caption{Support of least favorable prior $P_X^\star$ for $A=5$ and different values of $N$.}
		\label{fig:Position2}
	\end{subfigure}%
	~
	
	\begin{subfigure}[t]{.9\linewidth}
		\input{FIG/pdfA5N4.pgf}\vspace{-8pt}
		\caption{Least favorable priors for $A=5$ and $N=1,3,4$.}
		\label{fig:Prob2}
	\end{subfigure}%
	\caption{Least favorable priors for $P_{Y|X}$ given in \eqref{Eq:Gaussian_Case} as a result of quantization.}
	\label{fig:Gaussian}
\end{figure}
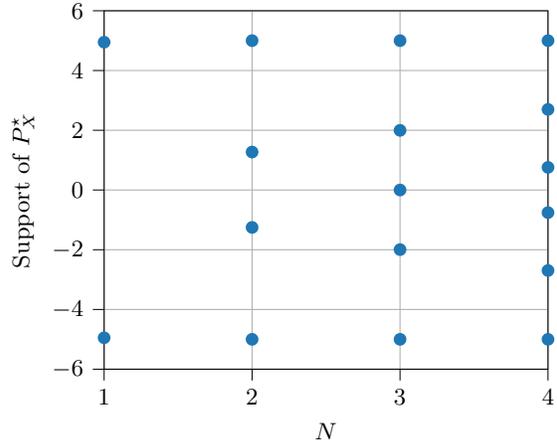
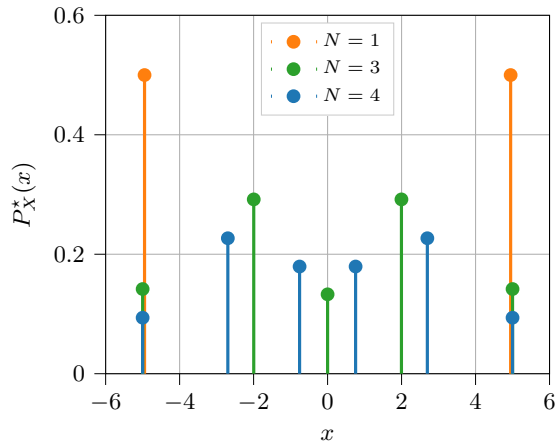

Now, for this model we would like to find the least favorable prior under the assumption that $X\in \Omega =[-A,A]$, $A>0$. The corresponding numerical results are depicted in Fig.~\ref{fig:Gaussian} for $A=5$ and different $N$.  
%
%
%
%%%%%%%%%%%%%%%%%%%%%%%%%%%%%%%%%%%%%%%%%%%%%%%%%%%%%%%%%%%%%%%%%%%%%%%%%%%%%%%%%%%%%%%%%%%%%%%%%%%%%%%%%%%%%%%%%%%%%%%%%%%
\section{CONCLUSION} 
This work has developed a dimensionality reduction method for finding least favorable prior distributions. The method produces an explicit bound on the size of the support of a least favorable prior and thus transforms the corresponding optimization problem from an infinite-dimensional to a finite-dimensional one. The numerical benefits of this method have been demonstrated via several examples based on a projected gradient ascent algorithm. Finally, while the focus was on loss functions that belong to the family of Bregman divergences, for which the optimal estimator is the conditional mean, the method can be generalized to other loss functions that have a concrete optimal estimator. For example, the results can be extended to the $L_1$ case, where the optimal Bayesian estimator is also the conditional median. 
\acknowledgments%
{We}\! would like to thank the anonymous reviewers for their comments and suggestions, which helped to improve the quality of this paper. The work of H.\,V. Poor was supported by the U.S. National Science Foundation (NSF) within the Israel-US Binational program under grant CCF-1908308 and the work of S. Shamai (Shitz) by the US-Israel Binational Science Foundation (BSF) under grant BSF-2018710, respectively.
\bibliography{refs}

\clearpage
\appendix

\thispagestyle{empty}

% For one-column format, uncomment the following:
%\onecolumn \makesupplementtitle
% For two-column format, uncomment the following:
\onecolumn \makesupplementtitle

%
%
%%%%%%%%%%%%%%%%%%%%%%%%%%%%%%%%%%%%%%%%%%%%%%%%%%%%%%%%%%%%%%%%%%%%%%%%%%%%%%%%%%%%%%%%%%%%%%%%%%%%%%%%%%%%%%%%%%%
\section{PROOF OF PROPOSITION~\ref{prop:Further_Reduction}}\label{sec:proofprop1}
Now that we know from Theorem~\ref{thm:BoundedOutPutChannel} that the least favorable distribution is discrete with at most $(k+1)(n+1)N$ mass points, we are able to slightly refine the number of mass points under various additional conditions. 
%
%
%
%%%%%%%%%%%%%%%%%%%%%%%%%%%%%%%%%%%%%%%%%%%%%%%%%%%%%%%%%%%%%%%%%%%%%%%%%%%%%%%%%%%%%%%%%%%%%%%%%%%%%%%%%%%%%%%%%%%
\subsection{Refinement for Compact $\Omega$}
Suppose that $\Omega$ is a proper compact subset of $\mathbb{R}^n$. Moreover, as the optimal input distribution exists, there exist numbers $c_m^\prime\le c_m$ such that
\begin{equation*}
\E_{P_X^\star}\bigl[ f_i(X)\bigr]=c^\prime_m\;,\;m=1,\dots,k.
\end{equation*}% 
Thus, the definition of the set in \eqref{eq:ConstraintSetfjalkfj;aldjflalkdfal} can be modified to
\begin{align*}
\mathcal{P}^\star=\Bigl\{P_X\in\mathcal{P}(\Omega):&\E_{P_X}\bigl[ X_j P_{Y|X_j}(y_i|X_j)\bigr] =c_{ji}^\star\,,\,P_{Y}(y_i; P_X)=P_{Y}(y_i; \star)\,,\,\E_{P_X}\bigl[f_i(X)\bigr]=c^\prime_m\\
&m=1,\dots,k\,,\,j=1,\dots,n\,,\,i=1,\dots,N\Bigr\}. 
\end{align*}%
Clearly, $P_X^\star\in\mathcal{P}^\star$. Note that $\mathcal{P}^\star$ is an intersection of the hyperplanes defined in  \eqref{eq:hyperplanceFromTheOutput}  and \eqref{eq:hyperplanceNumeratorOfConditionalExpectation} with the hyperplanes
\begin{equation*}
\mathcal{V}_i\coloneqq\bigl\{P_X\in\mathcal{P}(\Omega):\E_{P_X}\bigl[f_i(X)\bigr]=c_m^\prime\bigr\}\;,\;i=1,\dots,k.
\end{equation*}%
Moreover, each $\mathcal{V}_i$ is a closed set as $f_i$ is bounded and continuous on $\Omega$ for each $i=1,\dots,k$ (see Theorem~\ref{thm:ContinuityOfLinearFunctionals}). Therefore, applying Dubins' theorem to the set $\mathcal{P}^\star$, it follows that the extreme points of $\mathcal{P}^\star$ can be represented as convex combinations of at most $(n+1)(N-1)+k+1$ extreme points of $\mathcal{P}(\Omega)$. As the extreme points of $\mathcal{P}(\Omega)$ are point masses, we have that the least favorable distribution $P_X^\star$ has at most $(n+1)(N-1)+k+1$ mass points. 
%
%
%
%%%%%%%%%%%%%%%%%%%%%%%%%%%%%%%%%%%%%%%%%%%%%%%%%%%%%%%%%%%%%%%%%%%%%%%%%%%%%%%%%%%%%%%%%%%%%%%%%%%%%%%%%%%%%%%%%%%
\subsection{Refinement for Functions that Imply $P_{X_j}(+\infty)=P_{X_j}(-\infty)=0$ for all $1 \le j\le n$} 
Now, let $\Omega$ be arbitrary and $f_1,\dots,f_k$ bounded and continuous on $\Omega$ such that for every $P_X$ with a finite number of mass points the conditions $\E_{P_X}\bigl[f_i(X)\bigr]<\infty$, $i=1,\dots,k$, imply that $P_{X_j}(+\infty)=P_{X_j}(-\infty)=0$ for all $j=1,\dots,n$. Then, this implies that the least favorable distribution, which has at most $(k+1)(n+1)N$ point masses, must have a bounded support. Since the support is bounded, we can refine the bound on the number of mass points to $(n+1)(N-1)+k+1$.

This concludes the proof. 
%
%
%
%%%%%%%%%%%%%%%%%%%%%%%%%%%%%%%%%%%%%%%%%%%%%%%%%%%%%%%%%%%%%%%%%%%%%%%%%%%%%%%%%%%%%%%%%%%%%%%%%%%%%%%%%%%%%%%%%%%
\section{PROOF OF PROPOSITION~2}\label{sec:Grad_MMSE}
In this section, we present the detailed proof of Proposition~2. Choosing $\phi(x)=x^2$ generates the canonical example of a Bregman divergence, namely
\begin{equation}
	\ell_\phi\bigl(X,f(Y)\bigr)=\bigl(X-f(Y)\bigr)^2.
\end{equation}%
With part 5 of Theorem~1, the corresponding Bayesian risk is then of the form
\begin{equation}
	R_\phi(P_X,P_{Y|X})=\E\bigl[\ell_\phi\bigl(X,\E[X|Y]\bigr)\bigr]=\E\bigl[\bigl(X-\E[X|Y]\bigr)^2\bigr]\eqqcolon g(\mathbf{x}),
	\label{eq:g}
\end{equation}%
which is nothing but the minimum mean square error. In what follows, we assume that $g$ is differentiable.

%{\color{red}[MG] Now I've actually realized that Proposition 2 is referring to the special case $n=1$ only; that is, $\Omega\subseteq\mathbb{R}$. Hence, $\phi(x)=x^2$ and not $\phi(x)=\|x\|^2$ as stated in the proposition. Therefore, (2) is equal to (3).}

To compute the gradient of $g$, the following well-known formula will be useful: 
\begin{equation}
	\E\bigl[\bigl(X-\E[X|Y]\bigr)^2\bigr]= \E[X^2] - \E\bigl[\E[X|Y]^2\bigr]. 
	\label{eq:mmse}
\end{equation}%
%
%
%
%%%%%%%%%%%%%%%%%%%%%%%%%%%%%%%%%%%%%%%%%%%%%%%%%%%%%%%%%%%%%%%%%%%%%%%%%%%%%%%%%%%%%%%%%%%%%%%%%%%%%%%%%%%%%%%%%%%
\subsection{Partial Derivatives with Respect to Probabilities} 
In this subsection we focus on finding the partial derivatives of (\ref{eq:g}) with respect to the probability masses $p_i$, $i=1,\dots,N$. Towards this end, consider the first term on the right-hand side of (\ref{eq:mmse}) and observe that
\begin{align}
	\frac{\partial}{\partial p_i}\E[X^2]=\frac{\partial}{\partial p_i}\sum_{k=1}^N p_kx_k^2=x_i^2\;\,,\;\,i=1,\dots,N.
	\label{eq:first_term_der_p}
\end{align}%

In order to find the partial derivative of the second term (i.e., $\partial\,\E[\E^2[X|Y]]/\partial p_i $), note first that
\begin{equation}
	\frac{\partial}{\partial p_i} P_{Y}(y_j)=\frac{\partial}{\partial p_i}\sum_{k=1}^Np_k P_{Y|X} (y_j|x_k)= P_{Y|X} (y_j|x_i)\;,\;i=1,\dots,N.
	\label{eq:Derivative_of_p_of_output}
\end{equation}%
Furthermore, note that for $i=1,\dots,N$ 
\begin{align}
	\frac{\partial}{\partial p_i} \E[X| Y=y_j]&=\frac{\partial}{\partial p_i}\frac{\E\bigl[X P_{Y|X}(y_j|X)\bigr]}{P_{Y}(y_j)}\\[5pt]
	&=\frac{P_{Y}(y_j)\frac{\partial}{\partial p_i }\E\bigl[X P_{Y|X}(y_j|X)\bigr]-\E\bigl[X P_{Y|X}(y_j|X)\bigr]\frac{\partial}{\partial p_i}P_{Y}(y_j)}{\bigl(P_{Y}(y_j)\bigr)^2}\label{eq:usingDerivative_of_output}\\[5pt]
	&=\frac{P_{Y}(y_j)x_i P_{Y|X}(y_j|x_i)-\E\bigl[X P_{Y|X}(y_j|X)\bigr]P_{Y|X}(y_j|x_i)}{\bigl(P_{Y}(y_j)\bigr)^2}\\
	&=\frac{x_i P_{Y|X}(y_j|x_i)}{P_{Y}(y_j)}-\E[X|Y=y_j]\frac{P_{Y|X}(y_j|x_i)}{P_{Y}(y_j)}\\
	&=\bigl(x_i-\E[X|Y=y_j]\bigr)\frac{P_{Y|X}(y_j|x_i)}{P_{Y}(y_j)}, \label{eq:Derivative_of_Conditional_Expectation} 
\end{align}%
where in \eqref{eq:usingDerivative_of_output}  we have used the derivative in \eqref{eq:Derivative_of_p_of_output} and 
\begin{equation}
	\frac{\partial}{\partial p_i}\E\bigl[X P_{Y|X} (y_j|X)\bigr]=\frac{\partial}{\partial p_i}\sum_{k=1}^N p_kx_kP_{Y|X}(y_j|x_k )=x_i  P_{Y|X}(y_j|x_i). 
\end{equation}% 

Now,
{\small
\begin{align}
	\frac{\partial}{\partial p_i}\E\bigl[\E[X|Y]^2\bigr]&=\frac{\partial}{\partial p_i }\sum_{j=1}^N  P_{Y}(y_j)\E[X|Y=y_j]^2\\
	&=\sum_{j=1}^N\E[X|Y=y_j]^2\frac{\partial}{\partial p_i}P_{Y}(y_j)+\sum_{j=1}^NP_{Y}(y_j)\frac{\partial}{\partial p_i}\E[X|Y=y_j]^2\\
	&=\sum_{j=1}^N P_{Y|X}(y_j|x_i)\E[X|Y=y_j]^2+2\sum_{j=1}^N  P_{Y}(y_j)\E[X|Y=y_j]\bigl(x_i-\E[X|Y=y_j]\bigr)\frac{P_{Y|X}(y_j|x_i)}{P_{Y}(y_j)}\label{Eq:using_all_derivative_Expressions}\\[5pt]
	&= \E\bigl[\E[X|Y]^2\,\big|\,X=x_i\bigr] + 2\E\bigl[\E[X|Y](X- \E[X|Y])\,\big|\,X=x_i\bigr]\\[5pt]
	&=2 x_i\E\bigl[\E[X|Y]\,\big|\,X=x_i\bigr] -\E\bigl[\E[X|Y]^2\,\big|\,X=x_i], \label{eq:derivative_p_second_term}
\end{align}%
}%
where in \eqref{Eq:using_all_derivative_Expressions} we have used expressions \eqref{eq:Derivative_of_p_of_output} and \eqref{eq:Derivative_of_Conditional_Expectation}. 

Finally, combining \eqref{eq:first_term_der_p} and \eqref{eq:derivative_p_second_term} results in
\begin{align}
	\frac{\partial}{\partial p_i}\E\bigl[\bigl(X-\E[X|Y]\bigr)^2\bigr]&=x_i^2-2 x_i\E\bigl[\E[X|Y]\,\big|\,X=x_i\bigr]+\E\bigl[\E[X|Y]^2\,\big|\,X=x_i]\\
	&=\E\bigl[\bigr(x_i-\E[X|Y])^2\,\big|\,X=x_i],
\end{align}% 
$i=1,\dots,N$.
%
%
%
%%%%%%%%%%%%%%%%%%%%%%%%%%%%%%%%%%%%%%%%%%%%%%%%%%%%%%%%%%%%%%%%%%%%%%%%%%%%%%%%%%%%%%%%%%%%%%%%%%%%%%%%%%%%%%%%%%%
\subsection{Partial Derivatives with Respect to Locations}
Now, we focus on finding the partial derivatives of (\ref{eq:g}) with respect to the locations $x_i$, $i=1,\dots,N$, of the probability masses. Therefore, consider again the first term on the right-hand side of (\ref{eq:mmse}):
\begin{equation}
	\frac{\partial}{\partial x_i}\E[X^2]=\frac{\partial}{\partial x_i}\sum_{k=1}^Np_k x_k^2=2 x_i p_i\;\,,\;\,i=1,\dots,N.
\label{eq:derivative_of_the_frist_sterm_x} 
\end{equation}% 
To obtain the partial derivatives of the second term (\ref{eq:mmse}), first observe that
\begin{equation}
	\frac{\partial}{\partial x_i}P_{Y}(y_j)=\frac{\partial}{\partial x_i }\sum_{k=1}^N p_k  P_{Y|X} (y_j|x_k)= p_i\frac{\partial}{\partial x_i}P_{Y|X}(y_j|x_i)=p_i P_{Y|X}'(y_j|x_i).
	\label{eq:Derivative_output_position}
\end{equation}%
Furthermore, we need the partial derivatives of the conditional expectation with respect to the locations, which is given by 
\begin{align}
	\frac{\partial}{\partial x_i}\E[X|Y=y_j]&=\frac{P_{Y}(y_j)\frac{\partial}{\partial x_i}\E\bigl[X P_{Y|X}(y_j|X)\bigr]-\E\bigl[X P_{Y|X}(y_j|X)\bigr]\frac{\partial}{\partial x_i}P_{Y}(y_j)}{\bigl(P_{Y}(y_j)\bigr)^2}\\
	&=p_i\frac{x_i P_{Y|X}'(y_j|x_i)+P_{Y|X}(y_j|x_i)}{P_{Y}(y_j)}-\E[X|Y=y_j]\frac{p_i P_{Y|X}'(y_j|x_i)}{P_{Y}(y_j)} \label{eq:Applying_Derivatives} \\
	&=p_i\frac{\bigl(x_i-\E[X|Y=y_j]\bigr)P_{Y|X}'(y_j|x_i)+P_{Y|X}(y_j|x_i)}{P_{Y}(y_j)}, 
\end{align}%
where in \eqref{eq:Applying_Derivatives} we have used the partial derivative \eqref{eq:Derivative_output_position} together with
\begin{equation}
	\frac{\partial}{\partial x_i}\E\bigl[X P_{Y|X}(y_j|X)\bigr]=\frac{\partial}{\partial p_i}\sum_{k=1}^N p_k  x_k  P_{Y|X}(y_j|x_k)=p_ix_iP_{Y|X}'(y_j|x_i)+p_iP_{Y|X}(y_j|x_i). 
\end{equation}% 
Now, following along similar lines as in the previous subsection we obtain
{\small
\begin{align}
	\frac{\partial}{\partial x_i}\E\bigl[\E[X|Y]^2\bigr]&=\frac{\partial}{\partial x_i}\sum_{j=1}^N P_{Y}(y_j)\E[X| Y=y_j]^2\\
	&=\sum_{j=1}^N\E[X|Y=y_j]^2\frac{\partial}{\partial x_i}P_{Y}(y_j)+2\sum_{j=1}^NP_{Y}(y_j)\E[X|Y=y_j]\frac{\partial}{\partial x_i}\E[X| Y=y_j]\\
	&=\sum_{j=1}^Np_i P_{Y|X}'(y_j|x_i)\E[X|Y=y_j]^2+2p_i\sum_{j=1}^N\E[X|Y=y_j]\bigl(x_i-\E[X|Y=y_j]\bigr)P_{Y|X}'(y_j|x_i)+P_{Y|X} (y_j|x_i)\bigr)\\
	&=2 p_i  \sum_{j=1}^N    P_{Y|X} (y_j|x_i)  \E[X| Y=y_j]+ 2 p_i x_i  \sum_{j=1}^N    P_{Y|X}' (y_j|x_i)  \E[X| Y=y_j]\\
	&\hspace{141pt}-\sum_{j=1}^N    p_i P_{Y|X}' (y_j|x_i)  \E[X| Y=y_j]^2. 
	\label{eq:Derivative_of_the_second-term_x}
\end{align}%
}%

Finally, combining \eqref{eq:derivative_of_the_frist_sterm_x} and \eqref{eq:Derivative_of_the_second-term_x} results in
{\small
\begin{align}
	\frac{\partial}{\partial x_i}\E\bigl[\bigl(X-\E[X|Y]\bigr)^2\bigr]&=2 x_i p_i-2 p_i\sum_{j=1}^NP_{Y|X} (y_j|x_i)\E[X|Y=y_j]-2 p_i x_i  \sum_{j=1}^N    P_{Y|X}' (y_j|x_i)  \E[X| Y=y_j]\\
	&\hspace{174pt}+ \sum_{j=1}^N    p_i P_{Y|X}' (y_j|x_i)  \E[X| Y=y_j]^2\\
	&= 2 p_i \left (x_i -\sum_{j=1}^N    P_{Y|X} (y_j|x_i)  \E[X| Y=y_j]  \right)+p_i  \sum_{j=1}^N    P_{Y|X}' (y_j|x_i)  \bigl( \E[X| Y=y_j]^2\\
	&\hspace{176pt} -2 x_i   \E[X| Y=y_j]\bigr) \\
	&= 2 p_i \Bigl(x_i-\E\bigl[\E[X| Y]\,\big|\,X=x_i\bigr]\Bigr)+p_i  \sum_{j=1}^N    P_{Y|X}' (y_j|x_i)  \left( \E[X| Y=y_j]^2  -2 x_i   \E[X| Y=y_j]\right), 
\end{align}%
}%
which concludes the proof.

\end{document}

%% file: FIG/cube_plane.pgf
\begin{tikzpicture}[3d view={120}{20},line join=round,
 declare function={a=4;b=2;}]
 \draw[very thick, style=dashed, color=black] (a,0,-a) -- (0,0,-a)-- (0,a,-a);
 \draw[very thick, style=dashed, color=black] (0,0,0) -- (0,0,-a); 
 \draw[very thick,black] (a,0,0) -- (a,0,-a) -- (a,a,-a) -- (a,a,-b);
 \draw[very thick,black] (a,a-b,0) -- (a,0,0) -- (0,0,0) -- (0,a,0) -- (a-b,a,0);
 \draw[very thick,black] (0,a,0) -- (0,a,-a) -- (a,a,-a);
 %\fill[gray,opacity=0.6] (4,0.5,1.5) -- (0.5,4,1.5) -- (2.75,6.25,-3) -- (6.25,2.75,-3) -- cycle;
 \fill[gray,opacity=0.6] (4,1,1) -- (1,4,1) -- (2.75,5.75,-2.5) -- (5.75,2.75,-2.5) -- cycle;
 %\draw[very thick,pattern=north east lines] (a,a,-b) -- (a-b,a,0) -- (a,a-b,0) -- cycle;
 \draw[very thick,fill=white,opacity=0.9] (a,a,-b) -- (a-b,a,0) -- (a,a-b,0) -- cycle;  
 \draw[very thick,black] (a-b,a,0) -- (a,a,0) -- (a,a-b,0);
 \draw[very thick,black] (a,a,0) -- (a,a,-b);
 \node[circle,draw,fill=black,label=below left:$A$,scale=0.4] at (4,2,0){};
 \node[circle,draw,fill=black,label=below left:$B$,scale=0.4] at (4,4,0){};
 \node[circle,draw,fill=black,label=below left:$C$,scale=0.4] at (4,0,0){};
 \node[coordinate,pin={[pin distance=0.6cm]-45:{$\mathcal{I}$}}] at (5,5,0){};
\end{tikzpicture}

%% file: FIG/PositionBinomial.pgf
% This file was created by tikzplotlib v0.8.5.
\begin{tikzpicture}

\pgfplotsset{every tick label/.append style={font=\footnotesize}}
\pgfplotsset{every axis label/.append style={font=\footnotesize}}
\pgfplotsset{every legend label/.append style={font=\footnotesize}}
\pgfplotsset{legend style={font=\scriptsize}}

\definecolor{myblue}{RGB}{31,119,180}

\begin{axis}[
legend cell align={left},
legend style={draw=white!80.0!black},
tick align=outside,
tick pos=left,
x grid style={white!69.01960784313725!black},
xlabel={$m$},
width=2.946in,
height=2.5in,
xmajorgrids,
xmin=0, xmax=10,
xtick style={color=black},
y grid style={white!69.01960784313725!black},
ylabel={Support of $P_X^\star$},
ymajorgrids,
ymin=0, ymax=1,
ytick style={color=black}
]
\addplot [thick, myblue, mark=*, mark options={solid}, only marks]
table {%
1	0.146446609406727
1	0.853553390593277
2	0.0925568778832533
2	0.500000000000000
2	0.907443122116751
3	0.0774168358391555
3	0.354720642042881
3	0.645279357957129
3	0.922583164160836
4	0.0699538630648165
4	0.277669611364154
4	0.500000000000000
4	0.722330388635855
4	0.930046136935165
5	0.0606813430736565
5	0.234732985608626
5	0.399638437200498
5	0.600361562799501
5	0.765267014391374
5	0.939318656926344
6	0.0569014182443989
6	0.202714632247173
6	0.342780362151958
6	0.500000000000000
6	0.657219637848034
6	0.797285367752826
6	0.943098581755599
7	0.0529319550472015
7	0.177223190264948
7	0.300521683055523
7	0.433338680364827
7	0.566661319635172
7	0.699478316944474
7	0.822776809735055
7	0.947068044952800
8	0.0486460121352966
8	0.158126135429660
8	0.266109448317524
8	0.384527072329855
8	0.500000000000000
8	0.615472927670142
8	0.733890551682477
8	0.841873864570334
8	0.951353987864703
9	0.0447352081915576
9	0.142431809223023
9	0.239575472863018
9	0.344542085370812
9	0.448971276955860
9	0.551028723044142
9	0.655457914629186
9	0.760424527136981
9	0.857568190776969
9	0.955264791808439
10	0.0412031564181314
10	0.128599303678107
10	0.218388489231665
10	0.312394314378196
10	0.406558362492566
10	0.500000000000000
10	0.593441637507438
10	0.687605685621806
10	0.781611510768332
10	0.871400696321894
10	0.958796843581869
};
\end{axis}

\end{tikzpicture}

%% file: FIG/BinomialPMFm3610.pgf
% This file was created by tikzplotlib v0.8.5.
\begin{tikzpicture}

\pgfplotsset{every tick label/.append style={font=\footnotesize}}
\pgfplotsset{every axis label/.append style={font=\footnotesize}}
\pgfplotsset{every legend label/.append style={font=\footnotesize}}
\pgfplotsset{legend style={font=\scriptsize}}

\definecolor{myblue}{RGB}{31,119,180}
\definecolor{myred}{RGB}{255,127,14}
\definecolor{mygreen}{RGB}{44,160,44}

\begin{axis}[
legend cell align={left},
legend style={draw=white!80.0!black},
tick align=outside,
tick pos=left,
x grid style={white!69.01960784313725!black},
xlabel={$x$},
width=2.946in,
height=2.5in,
xmajorgrids,
xmin=0, xmax=1,
xtick style={color=black},
y grid style={white!69.01960784313725!black},
ylabel={$P_X^\star(x)$},
ymajorgrids,
ymin=0, ymax=0.4,
ytick style={color=black}
]
\addplot [ycomb, myblue, very thick, mark=*]
table {%
0.0774168358391555	0.223534841337280
0.354720642042881	0.276465158662724
0.645279357957129	0.276465158662715
0.922583164160836	0.223534841337282
};
\addlegendentry{$m=3$}

\addplot [ycomb, myred, very thick, mark=*]
table {%
0.0569014182443989	0.0967550439408599
0.202714632247173	0.147768146722433
0.342780362151958	0.169154914911649
0.500000000000000	0.172643788850112
0.657219637848034	0.169154914911651
0.797285367752826	0.147768146722433
0.943098581755599	0.0967550439408620
};
\addlegendentry{$m=6$}

\addplot [ycomb, mygreen, very thick, mark=*]
table {%
0.0412031564181314	0.0333049258479494
0.128599303678107	0.0744375761638176
0.218388489231665	0.0977752576483025
0.312394314378196	0.112107228817915
0.406558362492566	0.120625253626088
0.500000000000000	0.123499515791856
0.593441637507438	0.120625253626088
0.687605685621806	0.112107228817913
0.781611510768332	0.0977752576483015
0.871400696321894	0.0744375761638186
0.958796843581869	0.0333049258479495
};
\addlegendentry{$m=10$}
\end{axis}

\end{tikzpicture}

%% file: FIG/QuantFunction.pgf
% This file was created by tikzplotlib v0.8.5.
\begin{tikzpicture}

\pgfplotsset{every tick label/.append style={font=\footnotesize}}
\pgfplotsset{every axis label/.append style={font=\footnotesize}}
\pgfplotsset{every legend label/.append style={font=\footnotesize}}
\pgfplotsset{legend style={font=\scriptsize}}

\definecolor{myblue}{RGB}{31,119,180}
\definecolor{myred}{RGB}{255,127,14}
\definecolor{mygreen}{RGB}{44,160,44}

\begin{axis}[
legend cell align={left},
legend style={draw=white!80.0!black},
legend pos = north west,
tick align=outside,
tick pos=left,
x grid style={white!69.01960784313725!black},
xlabel={$w$},
width=2.946in,
height=2.5in,
xmajorgrids,
xmin=-8, xmax=8,
xtick style={color=black},
xtick={-8,-6,-4,-2,0,2,4,6,8},
y grid style={white!69.01960784313725!black},
ylabel={$Q_N(w)$},
ymajorgrids,
ymin=-3, ymax=3,
ytick style={color=black},
ytick={-3,-2,-1,0,1,2,3}
]
\addplot [myred, very thick]  coordinates {(-8,-1) (-0.5,-1) (-0.5,0) (0.5,0) (0.5,1) (8,1)};
\addlegendentry{$N=1$}
\addplot [mygreen, very thick]  coordinates {(-8,-2) (-1.5,-2) (-1.5,-1) (-0.5,-1) (-0.5,0) (0.5,0) (0.5,1) (1.5,1) (1.5,2) (8,2)};
\addlegendentry{$N=2$}
\addplot [myblue, very thick]  coordinates {(-8,-3) (-2.5,-3) (-2.5,-2) (-1.5,-2) (-1.5,-1) (-0.5,-1) (-0.5,0) (0.5,0) (0.5,1) (1.5,1) (1.5,2) (2.5,2) (2.5,3) (8,3)};
\addlegendentry{$N=3$}

\end{axis}

\end{tikzpicture}

%% file: FIG/Gn4A5.pgf
\begin{tikzpicture}

\pgfplotsset{every tick label/.append style={font=\footnotesize}}
\pgfplotsset{every axis label/.append style={font=\footnotesize}}
\pgfplotsset{every legend label/.append style={font=\footnotesize}}
\pgfplotsset{legend style={font=\scriptsize}}

\definecolor{myblue}{RGB}{31,119,180}

\begin{axis}[
legend cell align={left},
legend style={draw=white!80.0!black},
tick align=outside,
tick pos=left,
x grid style={white!69.01960784313725!black},
xlabel={$N$},
width=2.946in,
height=2.5in,
xmajorgrids,
xmin=1, xmax=4,
xtick style={color=black},
y grid style={white!69.01960784313725!black},
ylabel={Support of $P_X^\star$},
ymajorgrids,
ymin=-6, ymax=6,
ytick style={color=black},
ytick={-6,-4,-2,0,2,4,6}
]
\addplot [thick, myblue, mark=*, mark options={solid}, only marks]
table {%
4	5
4	2.69667732840234
4	0.757135665915334
4	-0.757135665915334
4	-2.69667732840234
4	-5
3	5
3	1.99630404462921
3	6.00508453516951e-17
3	-1.99630404462921
3	-5
2	5
2	1.27052270099780
2	-1.25352051093667
2	-5
1	4.94831173577928
1	-4.94831173577927
};
\end{axis}

\end{tikzpicture}

%% file: FIG/pdfA5N4.pgf
% This file was created by tikzplotlib v0.8.5.
\begin{tikzpicture}

\pgfplotsset{every tick label/.append style={font=\footnotesize}}
\pgfplotsset{every axis label/.append style={font=\footnotesize}}
\pgfplotsset{every legend label/.append style={font=\footnotesize}}
\pgfplotsset{legend style={font=\scriptsize}}

\definecolor{myblue}{RGB}{31,119,180}
\definecolor{myred}{RGB}{255,127,14}
\definecolor{mygreen}{RGB}{44,160,44}

\begin{axis}[
legend cell align={left},
legend style={draw=white!80.0!black},
legend style={at={(0.5,0.98)},anchor=north},
tick align=outside,
tick pos=left,
x grid style={white!69.01960784313725!black},
xlabel={$x$},
width=2.946in,
height=2.5in,
xmajorgrids,
xmin=-6, xmax=6,
xtick style={color=black},
y grid style={white!69.01960784313725!black},
ylabel={$P_X^\star(x)$},
ymajorgrids,
ymin=0, ymax=0.6,
ytick style={color=black}
]
\addplot [ycomb, myred, very thick, mark=*]
table {%
-4.94831173577927	0.499999999999997
4.94831173577928	0.500000000000003
};
\addlegendentry{$N=1$}

\addplot [ycomb, mygreen, very thick, mark=*]
table {%
5	0.141745421748132
1.99630404462921	0.291810770005432
6.00508453516951e-17	0.132887616492871
-1.99630404462921	0.291810770005432
-5	0.141745421748133
};
\addlegendentry{$N=3$}

\addplot [ycomb, myblue, very thick, mark=*]
table {%
5	0.0937570955637619
2.69667732840234	0.226795256369947
0.757135665915334	0.179447648066291
-0.757135665915334	0.179447648066291
-2.69667732840234	0.226795256369947
-5	0.0937570955637619
};
\addlegendentry{$N=4$}
\end{axis}

\end{tikzpicture}